\documentclass[journal]{IEEEtran}
\usepackage{bm}

\usepackage{amsfonts}
\usepackage{graphicx}
\usepackage{makecell}
\usepackage{tikz}
\usepackage{collcell}
\usepackage{colortbl}
\usepackage{pgfplots}
\usepackage{pgfplotstable}
\usepackage{multirow}
\usepackage{url}
\usepackage[all]{xy}
\usepackage{cite}
\usepackage{booktabs}
\usepackage{adjustbox}
\usepackage{tikz}
\usepackage{arydshln}
\usepackage{epstopdf}
\usetikzlibrary{arrows,positioning}
\usetikzlibrary{plotmarks}
\usetikzlibrary{calc}

\usetikzlibrary{patterns}
\usetikzlibrary{shapes.geometric}
\pgfdeclarelayer{bg}    
\pgfsetlayers{bg,main}

\newcommand{\Pixel}{h}
\newcommand{\Patch}{\mathcal{P}}
\newcommand{\Patchx}{p_x}
\newcommand{\Patchy}{p_y}
\newcommand{\Patchw}{p_w}
\newcommand{\NumberOfClasses}{\mathcal{C}}
\newcommand{\NumberOfKernels}{\mathcal{K}}

\newcommand{\NumberOfBands}{\mathcal{B}}

\newcommand{\term}[1]{\mathfrak{#1}}
\newcommand{\zbior}[1]{\mathbb{#1}}  
\newcommand{\macierz}[1]{{\boldsymbol{\mathrm{#1}}}}
\newcommand{\wektor}[1]{\macierz{\MakeLowercase{#1}}}
\newcommand{\newItem}[2]{%
  \expandafter\def\csname #1\endcsname {\MakeLowercase{#2}} %
  \expandafter\def\csname n#1\endcsname {\MakeUppercase{#2}} %
  \expandafter\def\csname zb#1\endcsname {\zbior{\MakeUppercase{#2}}} %
  \expandafter\def\csname t#1\endcsname {\term{\MakeLowercase{#2}}} %
  \expandafter\def\csname m#1\endcsname {\macierz{\MakeUppercase{#2}}} %
  \expandafter\def\csname w#1\endcsname {\wektor{#2}} %
  }
  \newItem{Atrybut}{d_f}
\newItem{Deskryptor}{x}
\newItem{Regula}{l}
\newItem{Przeslanka}{a}
\newItem{Konsekwencja}{b}
\newItem{Implikacja}{b'}
\newItem{Zespolona}{c}
\newItem{Rzeczywista}{r}
\newItem{Klasa}{\delta}
\newItem{Decyzja}{y}
\newItem{Przyklad}{a}
\newItem{Test}{a}
\newItem{Klaster}{c}
\newItem{Rozmycie}{s}
\newItem{Centre}{v}
\newItem{Membership}{u}
\newItem{ClusterMembership}{\mu}
\newItem{Weight}{z}
\newItem{Krotka}{x}
\newItem{Width}{w}
\newItem{Wspolczynnik}{p}
\newItem{Obiekt}{k}
\newItem{Odleglosc}{t}
\newItem{Parameter}{p}

\ifCLASSINFOpdf
\else
\fi

\pgfplotstableset{
    /color cells/min/.initial=0,
    /color cells/max/.initial=1000,
    /color cells/textcolor/.initial=,
    color cells/.code={%
        \pgfqkeys{/color cells}{#1}%
        \pgfkeysalso{%
            postproc cell content/.code={%
                \begingroup
                \pgfkeysgetvalue{/pgfplots/table/@preprocessed cell content}\value
\ifx\value\emptye
\endgroup
\else
                \pgfmathfloatparsenumber{\value}%
                \pgfmathfloattofixed{\pgfmathresult}%
                \let\value=\pgfmathresult
                \pgfplotscolormapaccess
                    [\pgfkeysvalueof{/color cells/min}:\pgfkeysvalueof{/color cells/max}]%
                    {\value}%
                    {\pgfkeysvalueof{/pgfplots/colormap name}}%
                \pgfkeysgetvalue{/pgfplots/table/@cell content}\typesetvalue
                \pgfkeysgetvalue{/color cells/textcolor}\textcolorvalue
                \toks0=\expandafter{\typesetvalue}%
                \xdef\temp{%
                    \noexpand\pgfkeysalso{%
                        @cell content={%
                            \noexpand\cellcolor[rgb]{\pgfmathresult}%
                            \noexpand\definecolor{mapped color}{rgb}{\pgfmathresult}%
                            \ifx\textcolorvalue\empty
                            \else
                                \noexpand\color{\textcolorvalue}%
                            \fi
                            \the\toks0 %
                        }%
                    }%
                }%
                \endgroup
                \temp
\fi
            }%
        }%
    }
}

\begin{document}
\title{Segmenting Hyperspectral Images Using Spectral-Spatial Convolutional Neural Networks With Training-Time Data Augmentation}
\author{Jakub Nalepa,~\IEEEmembership{Member,~IEEE}, 
        Lukasz Tulczyjew,
        Michal Myller,
        and Michal Kawulok,~\IEEEmembership{Member,~IEEE}
\thanks{This work was funded by European Space Agency (HYPERNET project).}
\thanks{J.~Nalepa, L.~Tulczyjew, M.~Myller, and M.~Kawulok are with Silesian University of Technology, Gliwice, Poland (e-mail: \{jnalepa, michal.kawulok\}@ieee.org), and KP Labs, Gliwice, Poland (\{jnalepa, ltulczyjew, mmyller, mkawulok\}@kplabs.pl).}
}


\maketitle
\begin{abstract}
Hyperspectral imaging provides detailed information about the scanned objects, as it captures their spectral characteristics within a large number of wavelength bands. Classification of such data has become an active research topic due to its wide applicability in a variety of fields. Deep learning has established the state of the art in the area, and it constitutes the current research mainstream. In this letter, we introduce a new spectral-spatial convolutional neural network, benefitting from a battery of data augmentation techniques which help deal with a real-life problem of lacking ground-truth training data. Our rigorous experiments showed that the proposed method outperforms other spectral-spatial techniques from the literature, and delivers precise hyperspectral classification in real time.
\end{abstract}


\IEEEpeerreviewmaketitle

\section{Introduction} \label{sec:intro}

Hyperspectral image analysis has been continuously gaining research attention, as the current advances in sensor design have made the acquisition of such imagery much more affordable nowadays. Hyperspectral images (HSI) capture a wide spectrum of light per pixel, and can be represented as large 3D tensors, commonly including hundreds of reflectance values per pixel, characterizing an underlying material. Classification and segmentation (assigning class labels to specific pixels, and finding the boundaries of the objects within the image, respectively) of HSI allow us to build an understanding of the captured scene. Such non-invasive extraction of the spectral footprint of the analyzed objects has multiple real-life applications ranging across various fields, including remote sensing, medicine, biochemistry, and more~\cite{8314827}.

In general, the approaches towards automatic HSI classification are divided into conventional image-analysis and machine-learning techniques, and deep learning-powered algorithms. In the former approaches, the input data undergoes the feature engineering process, in which hand-crafted features are extracted and selected~\cite{Bilgin2011,Dundar2018}. On the other hand, deep learning allows us to intrinsically learn an embedded data representation using the available training data~\cite{Chen2015,Zhao2016,Zhong2017,Mou2017,Santara2017,Lee2017,Gao_2018}. Deep learning has been tremendously successful in practically all fields related to computer vision and image analysis, and it is the current research mainstream in HSI analysis. Deep networks for HSI classification can be split into three groups, encompassing those operating exclusively on (i)~spectral and (ii)~spatial pixel characteristics, and (iii)~coupling spectral and spatial information for better performance~\cite{8697135,8340197}.

\begin{figure*}[ht!]
\hspace*{-0.7cm}
\resizebox{\textwidth}{!}{

\newcommand{\largefig}{120pt}
\newcommand{\mediumfig}{60pt}
\newcommand{\inputfig}{30pt}

\newcommand{\resnettext}{Reconstruction with ResNet}
\newcommand{\registrationtext}{Image registration}
\newcommand{\shifttext}{Shifts $(x,y)$}
\newcommand{\evoimtext}{EvoIM iterative image filtering}
\newcommand{\groundtruthtext}{$M$ ground-truth high-resolution images $\mathcal{I}^{(h)}$}

\newcommand{\blocksep}{45pt}
\newcommand{\multiblock}{5pt}
\newcommand{\arrowsep}{15pt}

\begin{tikzpicture}[scale=0.8,
        image/.style={inner sep=0pt,draw=white,very thick},
        whitefont/.style={text=white,font=\footnotesize},
        input/.style={rectangle,draw=black,fill=gray!40,inner sep=5pt,minimum height=27pt,minimum width=40pt,text width=50pt,text badly centered,font=\small,thick},
        output/.style={rectangle,draw=black,fill=red!20,inner sep=5pt,minimum height=32pt,minimum width=40pt,text width=65pt,text badly centered,font=\footnotesize,thick},
        ga/.style={rectangle,draw=black,fill=red!20,inner sep=5pt,minimum height=32pt,minimum width=40pt,text width=100pt,text badly centered,font=\footnotesize,thick,rounded
        corners=16pt},
        algstep/.style={rectangle,draw=black,fill=blue!20,inner sep=5pt,minimum height=27pt,minimum width=40pt,text width=60pt,text badly centered,font=\small,thick,rounded
        corners=8pt},
        plain/.style={rectangle,text width=40pt,text badly centered,font=\small},
        myarrow/.style={thick},
        external/.style={thick},
        internal/.style={dashed}]

\newcommand{\viewx}{0.4}
\newcommand{\viewy}{0.2}
\newcommand{\mysize}{5}
\newcommand{\mysizefilter}{0.7}
\newcommand{\slicedist}{0.15}
\newcommand{\slicenum}{10}
\newcommand{\coordsize}{1}
\newcommand{\mysizepoint}{0.2}

\coordinate (origin) at (0, 0);
\coordinate (pointer) at (origin);

\foreach \xx in {1,...,\slicenum}
{
    \coordinate (pointer) at ($(pointer) + (\slicedist, 0)$);

    \coordinate (A1\xx) at (pointer);
    \coordinate (A2\xx) at ($(A1\xx) + (0, \mysize) $);
    \coordinate (A3\xx) at ($(A2\xx) + (\viewx * \mysize, 0.7*\viewy * \mysize) $);
    \coordinate (A4\xx) at ($(A1\xx) + (\viewx * \mysize, \viewy * \mysize) $);

    \fill[gray!30] (A1\xx) -- (A2\xx) -- (A3\xx) -- (A4\xx) -- cycle;
    \draw[external] (A1\xx) -- (A2\xx)-- (A3\xx) -- (A4\xx) -- cycle;
}

\coordinate (pointer) at (origin);


    \draw[->,myarrow] ($(A12) + (0,-0.4)$) -- node[below] {$\NumberOfBands$} ($(A19) + (0,-0.4)$);
    \draw[->,myarrow] ($(A1\slicenum) + (0.6 + 0.25 * \viewx * \mysize, 0.25 * \viewy * \mysize)$) -- node[below right, inner sep=2] {$\Patchx$} ($(A1\slicenum) + (0.6 + 0.55 * \viewx * \mysize, 0.55 * \viewy * \mysize)$);
    \draw[->,myarrow] ($(A4\slicenum) + (0.2, 0.65 * \mysize)$) -- node[right,inner sep=2] {$\Patchy$} ($(A4\slicenum) + (0.2, 0.65 * \mysize + \coordsize)$);

\foreach \xxx in {1,2,...,\slicenum}
{
    \coordinate (Patch1\xxx) at ($(A1\xxx) + (0.5 * \viewx * \mysize, 0.3 * \mysize + 0.5 * \viewy * \mysize) - (0.33 * \mysizefilter,0.33*\mysizefilter) $);
    \coordinate (Patch2\xxx) at ($(Patch1\xxx) + (0, 1.67*\mysizefilter) $);
    \coordinate (Patch3\xxx) at ($(Patch2\xxx) + (1.67*\viewx * \mysizefilter, 1.67*0.9*\viewy * \mysizefilter) $);
    \coordinate (Patch4\xxx) at ($(Patch1\xxx) + (1.67*\viewx * \mysizefilter, 1.67*\viewy * \mysizefilter) $);

}

\foreach \xxx in {7,8,9}
{
    \coordinate (AI1\xxx) at ($(A1\xxx) + (0.5 * \viewx * \mysize, 0.2 * \mysize + 0.5 * \viewy * \mysize) $);
    \coordinate (AI2\xxx) at ($(AI1\xxx) + (0, \mysizefilter) $);
    \coordinate (AI3\xxx) at ($(AI2\xxx) + (\viewx * \mysizefilter, 0.9*\viewy * \mysizefilter) $);
    \coordinate (AI4\xxx) at ($(AI1\xxx) + (\viewx * \mysizefilter, \viewy * \mysizefilter) $);

    \fill[gray!90] (AI1\xxx) -- (AI2\xxx) -- (AI3\xxx) -- (AI4\xxx) -- cycle;
    \draw[external] (AI1\xxx) -- (AI2\xxx)-- (AI3\xxx) -- (AI4\xxx) -- cycle;
}
\foreach \xxx in {7,8,9}
{
     \draw[internal] (AI1\xxx) -- (AI2\xxx)-- (AI3\xxx) -- (AI4\xxx) -- cycle;
}

\foreach \xxx in {10,...,\slicenum}
{
    \coordinate (Patch1\xxx) at ($(A1\xxx) + (0.5 * \viewx * \mysize, 0.3 * \mysize + 0.5 * \viewy * \mysize) - (0.33 * \mysizefilter,0.33*\mysizefilter) $);
    \coordinate (Patch2\xxx) at ($(Patch1\xxx) + (0, 1.67*\mysizefilter) $);
    \coordinate (Patch3\xxx) at ($(Patch2\xxx) + (1.67*\viewx * \mysizefilter, 1.67*0.9*\viewy * \mysizefilter) $);
    \coordinate (Patch4\xxx) at ($(Patch1\xxx) + (1.67*\viewx * \mysizefilter, 1.67*\viewy * \mysizefilter) $);

}

\node at (AI18) [anchor=north] {\scriptsize 3};
\node at ($0.5*(AI27) + 0.5*(AI37) $) [anchor=south east, inner sep=0.3] {\scriptsize 3};
\node at ($0.5*(AI17) + 0.5*(AI27) $) [anchor=east, inner sep=0.5] {\scriptsize 3};

\foreach \xx in {1,...,\slicenum}
{
    \coordinate (pointer) at ($(pointer) + (\slicedist, 0)$);

    \coordinate (A1\xx) at (pointer);
    \coordinate (A2\xx) at ($(A1\xx) + (0, \mysize) $);
    \coordinate (A3\xx) at ($(A2\xx) + (\viewx * \mysize, 0.7*\viewy * \mysize) $);
    \coordinate (A4\xx) at ($(A1\xx) + (\viewx * \mysize, \viewy * \mysize) $);

    \draw[external] (A1\xx) -- (A2\xx);
}


\renewcommand{\mysize}{3.5}
\renewcommand{\slicenum}{10}
\newcommand{\boxdist}{3.0}

\coordinate (neworigin) at ($ 0.5*(A110) + 0.5*(A210) + (\boxdist,-0.5*\mysize)$);

\coordinate (pointer) at (neworigin);

    \coordinate (B1) at (pointer);
    \coordinate (B2) at ($(B1) + (0, \mysize) $);
    \coordinate (B3) at ($(B2) + (\viewx * \mysize, 0.7*\viewy * \mysize) $);
    \coordinate (B4) at ($(B1) + (\viewx * \mysize, \viewy * \mysize) $);
    \coordinate (B5) at ($ (B1) + (1*\mysize, 0)$);
    \coordinate (B6) at ($(B5) + (0, \mysize) $);
    \coordinate (B7) at ($(B6) + (\viewx * \mysize, 0.7*\viewy * \mysize) $);
    \coordinate (B8) at ($(B5) + (\viewx * \mysize, \viewy * \mysize) $);

    \fill[gray!30] (B5) -- (B1) -- (B2) -- (B3) -- (B7) -- (B8)  -- cycle;

    \draw[internal] (B1) -- (B4)-- (B3);
    \draw[internal] (B8)-- (B4);

    \draw[->,myarrow] ($0.5*(B1) + 0.5*(B5) + (-0.5*\coordsize,-0.4)$) -- node[below] {$(\NumberOfBands-2)$} ($0.5*(B1) + 0.5*(B5) + (0.5*\coordsize,-0.4)$);
    \draw[->,myarrow] ($(B5) + (0.6 + 0.15 * \viewx * \mysize, 0.15 * \viewy * \mysize)$) -- node[below right, inner sep=2] {$(\Patchx-2)$} ($(B5) + (0.6 + 0.55 * \viewx * \mysize, 0.55 * \viewy * \mysize)$);
    \draw[->,myarrow] ($(B8) + (0.2, 0.1 * \mysize)$) -- node[right,inner sep=2] {$(\Patchy-2)$} ($(B8) + (0.2, 0.1 * \mysize + \coordsize)$);

    \node at ($0.5*(B3) + 0.5*(B7)$) [anchor=south] {$24\times$};

    \coordinate (BI1) at ($(B1) + (0.33*\mysize, 0.6*\mysize) $);
    \coordinate (BI2) at ($(BI1) + (0, \mysizefilter) $);
    \coordinate (BI3) at ($(BI2) + (\viewx * \mysizefilter, 0.9*\viewy * \mysizefilter) $);
    \coordinate (BI4) at ($(BI1) + (\viewx * \mysizefilter, \viewy * \mysizefilter) $);
    \coordinate (BI5) at ($(BI1) + (\mysizefilter,0) $);
    \coordinate (BI6) at ($(BI5) + (0, \mysizefilter) $);
    \coordinate (BI7) at ($(BI6) + (\viewx * \mysizefilter, 0.9*\viewy * \mysizefilter) $);
    \coordinate (BI8) at ($(BI5) + (\viewx * \mysizefilter, \viewy * \mysizefilter) $);

    \fill[gray!90] (BI5) -- (BI1) -- (BI2) -- (BI3) -- (BI7) -- (BI8)  -- cycle;

    \draw[external] (BI1) -- (BI2) -- (BI3) -- (BI7) -- (BI8) -- (BI5) -- cycle;
    \draw[external] (BI5) -- (BI6) -- (BI7) -- (BI8) -- cycle;
    \draw[external] (BI1) -- (BI2) -- (BI6) -- (BI5) -- cycle;

    \draw[internal] (BI1) -- (BI4)-- (BI3);
    \draw[internal] (BI8)-- (BI4);

    \draw[external] (B1) -- (B2) -- (B3) -- (B7) -- (B8) -- (B5) -- cycle;
    \draw[external] (B5) -- (B6) -- (B7) -- (B8) -- cycle;
    \draw[external] (B1) -- (B2) -- (B6) -- (B5) -- cycle;

\node at ($0.5*(BI1) + 0.5*(BI5)$) [anchor=north] {\scriptsize 3};
\node at ($0.5*(BI2) + 0.5*(BI3) $) [anchor=south east, inner sep=0.3] {\scriptsize 3};
\node at ($0.5*(BI1) + 0.5*(BI2) $) [anchor=east, inner sep=0.5] {\scriptsize 3};

    \coordinate (point) at ($0.5*(AI19) + 0.5*(AI29) + (\boxdist + 0.6*\mysize, 0.0*\mysize)$);

    \coordinate (P1) at ($(point) + (0,0) $);
    \coordinate (P2) at ($(P1) + (0, \mysizepoint) $);
    \coordinate (P3) at ($(P2) + (\viewx * \mysizepoint, \viewy * \mysizepoint) $);
    \coordinate (P4) at ($(P1) + (\viewx * \mysizepoint, \viewy * \mysizepoint) $);
    \coordinate (P5) at ($(P1) + (\mysizepoint,0) $);
    \coordinate (P6) at ($(P5) + (0, \mysizepoint) $);
    \coordinate (P7) at ($(P6) + (\viewx * \mysizepoint, \viewy * \mysizepoint) $);
    \coordinate (P8) at ($(P5) + (\viewx * \mysizepoint, \viewy * \mysizepoint) $);

    \fill[black] (P1) -- (P2) -- (P3) -- (P7) -- (P8) -- (P5)  -- cycle;
    \draw[myarrow,dotted] (AI19) -- (P1);
    \draw[myarrow,dotted] (AI29) -- (P2);
    \draw[myarrow,dotted] (AI39) -- (P3);
    \draw[myarrow,dotted] (AI49) -- (P4);

%


\renewcommand{\mysize}{3.3}
\renewcommand{\slicenum}{10}
\renewcommand{\boxdist}{3.5}

    \coordinate (point) at ($0.5*(BI1) + 0.5*(BI2) + (\boxdist + 1.0*\mysize,0)$);

\coordinate (neworigin) at ($ 0.5*(B5) + 0.5*(B6) + (\boxdist,-0.5*\mysize)$);

\coordinate (pointer) at (neworigin);

    \coordinate (B1) at (pointer);
    \coordinate (B2) at ($(B1) + (0, \mysize) $);
    \coordinate (B3) at ($(B2) + (\viewx * \mysize, 0.7*\viewy * \mysize) $);
    \coordinate (B4) at ($(B1) + (\viewx * \mysize, \viewy * \mysize) $);
    \coordinate (B5) at ($ (B1) + (1.0*\mysize, 0)$);
    \coordinate (B6) at ($(B5) + (0, \mysize) $);
    \coordinate (B7) at ($(B6) + (\viewx * \mysize, 0.7*\viewy * \mysize) $);
    \coordinate (B8) at ($(B5) + (\viewx * \mysize, \viewy * \mysize) $);

    \fill[gray!30] (B5) -- (B1) -- (B2) -- (B3) -- (B7) -- (B8)  -- cycle;

    \draw[internal] (B1) -- (B4)-- (B3);
    \draw[internal] (B8)-- (B4);

    \draw[->,myarrow] ($0.5*(B1) + 0.5*(B5) + (-0.5*\coordsize,-0.4)$) -- node[below] {$(\NumberOfBands-4)$} ($0.5*(B1) + 0.5*(B5) + (0.5*\coordsize,-0.4)$);
    \draw[->,myarrow] ($(B5) + (0.6 + 0.15 * \viewx * \mysize, 0.15 * \viewy * \mysize)$) -- node[below right, inner sep=2] {$(\Patchx-4)$} ($(B5) + (0.6 + 0.55 * \viewx * \mysize, 0.55 * \viewy * \mysize)$);
    \draw[->,myarrow] ($(B8) + (0.2, 0.4 * \mysize)$) -- node[right,inner sep=2] {$(\Patchy-4)$} ($(B8) + (0.2, 0.4 * \mysize + \coordsize)$);

    \node at ($0.5*(B3) + 0.5*(B7)$) [anchor=south] {$24\times$};

%

    \coordinate (P1) at ($(point) + (0,0) $);
    \coordinate (P2) at ($(P1) + (0, \mysizepoint) $);
    \coordinate (P3) at ($(P2) + (\viewx * \mysizepoint, \viewy * \mysizepoint) $);
    \coordinate (P4) at ($(P1) + (\viewx * \mysizepoint, \viewy * \mysizepoint) $);
    \coordinate (P5) at ($(P1) + (\mysizepoint,0) $);
    \coordinate (P6) at ($(P5) + (0, \mysizepoint) $);
    \coordinate (P7) at ($(P6) + (\viewx * \mysizepoint, \viewy * \mysizepoint) $);
    \coordinate (P8) at ($(P5) + (\viewx * \mysizepoint, \viewy * \mysizepoint) $);

    \fill[black] (P1) -- (P2) -- (P3) -- (P7) -- (P8) -- (P5)  -- cycle;
    \draw[myarrow,dotted] (BI5) -- (P1);
    \draw[myarrow,dotted] (BI6) -- (P2);
    \draw[myarrow,dotted] (BI7) -- (P3);
    \draw[myarrow,dotted] (BI8) -- (P4);

    \coordinate (BI1) at ($(B1) + (0.5*\mysize, 0.35*\mysize) $);
    \coordinate (BI2) at ($(BI1) + (0, \mysizefilter) $);
    \coordinate (BI3) at ($(BI2) + (\viewx * \mysizefilter, 0.9*\viewy * \mysizefilter) $);
    \coordinate (BI4) at ($(BI1) + (\viewx * \mysizefilter, \viewy * \mysizefilter) $);
    \coordinate (BI5) at ($(BI1) + (\mysizefilter,0) $);
    \coordinate (BI6) at ($(BI5) + (0, \mysizefilter) $);
    \coordinate (BI7) at ($(BI6) + (\viewx * \mysizefilter, 0.9*\viewy * \mysizefilter) $);
    \coordinate (BI8) at ($(BI5) + (\viewx * \mysizefilter, \viewy * \mysizefilter) $);

    \fill[gray!90] (BI5) -- (BI1) -- (BI2) -- (BI3) -- (BI7) -- (BI8)  -- cycle;

    \draw[external] (BI1) -- (BI2) -- (BI3) -- (BI7) -- (BI8) -- (BI5) -- cycle;
    \draw[external] (BI5) -- (BI6) -- (BI7) -- (BI8) -- cycle;
    \draw[external] (BI1) -- (BI2) -- (BI6) -- (BI5) -- cycle;

    \draw[internal] (BI1) -- (BI4)-- (BI3);
    \draw[internal] (BI8)-- (BI4);

    \draw[external] (B1) -- (B2) -- (B3) -- (B7) -- (B8) -- (B5) -- cycle;
    \draw[external] (B5) -- (B6) -- (B7) -- (B8) -- cycle;
    \draw[external] (B1) -- (B2) -- (B6) -- (B5) -- cycle;

\node at ($0.5*(BI1) + 0.5*(BI5)$) [anchor=north] {\scriptsize 3};
\node at ($0.5*(BI2) + 0.5*(BI3) $) [anchor=south east, inner sep=0.3] {\scriptsize 3};
\node at ($0.5*(BI1) + 0.5*(BI2) $) [anchor=east, inner sep=0.5] {\scriptsize 3};


\renewcommand{\mysize}{3.1}
\renewcommand{\slicenum}{10}
\renewcommand{\boxdist}{3.5}

    \coordinate (point) at ($0.5*(BI1) + 0.5*(BI2) + (\boxdist + 1.2*\mysize,0)$);

\coordinate (neworigin) at ($ 0.5*(B5) + 0.5*(B6) + (\boxdist,-0.5*\mysize)$);

\coordinate (pointer) at (neworigin);

    \coordinate (B1) at (pointer);
    \coordinate (B2) at ($(B1) + (0, \mysize) $);
    \coordinate (B3) at ($(B2) + (\viewx * \mysize, 0.7*\viewy * \mysize) $);
    \coordinate (B4) at ($(B1) + (\viewx * \mysize, \viewy * \mysize) $);
    \coordinate (B5) at ($ (B1) + (1.0*\mysize, 0)$);
    \coordinate (B6) at ($(B5) + (0, \mysize) $);
    \coordinate (B7) at ($(B6) + (\viewx * \mysize, 0.7*\viewy * \mysize) $);
    \coordinate (B8) at ($(B5) + (\viewx * \mysize, \viewy * \mysize) $);

    \fill[gray!30] (B5) -- (B1) -- (B2) -- (B3) -- (B7) -- (B8)  -- cycle;

    \draw[internal] (B1) -- (B4)-- (B3);
    \draw[internal] (B8)-- (B4);

    \draw[->,myarrow] ($0.5*(B1) + 0.5*(B5) + (-0.5*\coordsize,-0.4)$) -- node[below] {$(\NumberOfBands-6)$} ($0.5*(B1) + 0.5*(B5) + (0.5*\coordsize,-0.4)$);
    \draw[->,myarrow] ($(B1) + (0.6 + 0.25 * \viewx * \mysize, 0.25 * \viewy * \mysize)$) -- node[right=3pt, inner sep=2] {$(\Patchx-6)$} ($(B1) + (0.6 + 0.65 * \viewx * \mysize, 0.65 * \viewy * \mysize)$);
    \draw[->,myarrow] ($(B4) + (0.2, 0.4 * \mysize)$) -- node[right,inner sep=2] {$(\Patchy-6)$} ($(B4) + (0.2, 0.4 * \mysize + \coordsize)$);

    \node at ($0.5*(B3) + 0.5*(B7)$) [anchor=south] {$24\times$};

    \coordinate (P1) at ($(point) + (0,0) $);
    \coordinate (P2) at ($(P1) + (0, \mysizepoint) $);
    \coordinate (P3) at ($(P2) + (\viewx * \mysizepoint, \viewy * \mysizepoint) $);
    \coordinate (P4) at ($(P1) + (\viewx * \mysizepoint, \viewy * \mysizepoint) $);
    \coordinate (P5) at ($(P1) + (\mysizepoint,0) $);
    \coordinate (P6) at ($(P5) + (0, \mysizepoint) $);
    \coordinate (P7) at ($(P6) + (\viewx * \mysizepoint, \viewy * \mysizepoint) $);
    \coordinate (P8) at ($(P5) + (\viewx * \mysizepoint, \viewy * \mysizepoint) $);

    \fill[black] (P1) -- (P2) -- (P3) -- (P7) -- (P8) -- (P5)  -- cycle;
    \draw[myarrow,dotted] (BI5) -- (P1);
    \draw[myarrow,dotted] (BI6) -- (P2);
    \draw[myarrow,dotted] (BI7) -- (P3);
    \draw[myarrow,dotted] (BI8) -- (P4);

    \draw[external] (B1) -- (B2) -- (B3) -- (B7) -- (B8) -- (B5) -- cycle;
    \draw[external] (B5) -- (B6) -- (B7) -- (B8) -- cycle;
    \draw[external] (B1) -- (B2) -- (B6) -- (B5) -- cycle;


\renewcommand{\mysize}{4}
\renewcommand{\slicenum}{10}
\renewcommand{\boxdist}{3}

\newcommand{\FCLayer}{1}
\newcommand{\FCLayerPrev}{0}

\coordinate (neworigin) at ($ 0.5*(B5) + 0.5*(B6) + (\boxdist, 1 + 0.7*\mysize)$);

\coordinate (pointer) at (neworigin);

\foreach \xx in {1, 2, 6}
{
    \node[circle,color=black, fill=black, inner sep=0pt,minimum size=7pt,label=below right:{}] (FC\FCLayer\xx) at ($(pointer) + (0, -\xx)$) {};

    \draw[->,myarrow,dotted] (B6) -- (FC\FCLayer\xx);
    \draw[->,myarrow,dotted] (B7) -- (FC\FCLayer\xx);
    \draw[->,myarrow,dotted] (B5) -- (FC\FCLayer\xx);
    \draw[->,myarrow,dotted] (B8) -- (FC\FCLayer\xx);

}

\coordinate (pointer) at ($0.5*(FC\FCLayer2) + 0.5*(FC\FCLayer6) + (0, 0.3)$);

\foreach \xx in { 0, 0.3, 0.6}
{
    \node[circle,color=black, fill=black, inner sep=0pt,minimum size=2pt,label=below right:{}] (b) at ($(pointer) + (0, -\xx)$) {};
}

\node (arrow) at (FC\FCLayer1.north) [anchor=south] {$\downarrow$};
\node at (arrow.east) [anchor=west, inner sep=0] {512 neurons};

%


\renewcommand{\mysize}{4}
\renewcommand{\slicenum}{10}
\renewcommand{\boxdist}{2}

\renewcommand{\FCLayer}{2}
\renewcommand{\FCLayerPrev}{1}

\coordinate (neworigin) at ($ (neworigin) + (\boxdist, -0.5)$);

\coordinate (pointer) at (neworigin);

\foreach \xx in {1, 2, 5}
{
    \node[circle,color=black, fill=black, inner sep=0pt,minimum size=7pt,label=below right:{}] (FC\FCLayer\xx) at ($(pointer) + (0, -\xx)$) {};

    \draw[->,myarrow,dotted] (FC\FCLayerPrev1) -- (FC\FCLayer\xx);
    \draw[->,myarrow,dotted] (FC\FCLayerPrev2) -- (FC\FCLayer\xx);
    \draw[->,myarrow,dotted] (FC\FCLayerPrev6) -- (FC\FCLayer\xx);
}

\coordinate (pointer) at ($0.5*(FC\FCLayer2) + 0.5*(FC\FCLayer5) + (0, 0.3)$);

\foreach \xx in { 0, 0.3, 0.6}
{
    \node[circle,color=black, fill=black, inner sep=0pt,minimum size=2pt,label=below right:{}] (b) at ($(pointer) + (0, -\xx)$) {};
}

\node (arrow) at (FC\FCLayer1.north) [anchor=south] {$\downarrow$};
\node at (arrow.east) [anchor=west, inner sep=0] {256 neurons};


\renewcommand{\mysize}{4}
\renewcommand{\slicenum}{10}
\renewcommand{\boxdist}{2}

\renewcommand{\FCLayer}{3}
\renewcommand{\FCLayerPrev}{2}

\coordinate (neworigin) at ($ (neworigin) + (\boxdist, -0.5)$);

\coordinate (pointer) at (neworigin);

\foreach \xx in {1, 2, 4}
{
    \node[circle,color=black, fill=black, inner sep=0pt,minimum size=7pt,label=below right:{}] (FC\FCLayer\xx) at ($(pointer) + (0, -\xx)$) {};

    \draw[->,myarrow,dotted] (FC\FCLayerPrev1) -- (FC\FCLayer\xx);
    \draw[->,myarrow,dotted] (FC\FCLayerPrev2) -- (FC\FCLayer\xx);
    \draw[->,myarrow,dotted] (FC\FCLayerPrev5) -- (FC\FCLayer\xx);
}

\coordinate (pointer) at ($0.5*(FC\FCLayer2) + 0.5*(FC\FCLayer4) + (0, 0.3)$);

\foreach \xx in { 0, 0.3, 0.6}
{
    \node[circle,color=black, fill=black, inner sep=0pt,minimum size=2pt,label=below right:{}] (b) at ($(pointer) + (0, -\xx)$) {};
}

\node (arrow) at (FC\FCLayer1.north) [anchor=south] {$\downarrow$};
\node at (arrow.east) [anchor=west, inner sep=0] {128 neurons};

%
%
%
%


\renewcommand{\mysize}{3}
\renewcommand{\slicenum}{15}
\renewcommand{\boxdist}{1.5}

\coordinate (neworigin) at ($ (neworigin) + (\boxdist, -0.5)$);

\coordinate (pointer) at (neworigin);

\foreach \xx in {1, 3}
{
    \node[circle,color=black, fill=black, inner sep=0pt,minimum size=7pt,label=below right:{}] (Out\xx) at ($(pointer) + (0, -\xx)$) {};

    \draw[->,myarrow,dotted] (FC\FCLayer1) -- (Out\xx);
    \draw[->,myarrow,dotted] (FC\FCLayer2) -- (Out\xx);
    \draw[->,myarrow,dotted] (FC\FCLayer4) -- (Out\xx);
}

    \draw[->,myarrow] (Out1) -- node[above, inner sep=2] {$O_1$} ($(Out1) + (1,0)$);
    \draw[->,myarrow] (Out3) -- node[above, inner sep=2] {$O_\NumberOfClasses$} ($(Out3) + (1,0)$);

\coordinate (pointer) at ($0.5*(Out1) + 0.5*(Out3) + (0, 0.3)$);

\foreach \xx in { 0, 0.3, 0.6}
{
    \node[circle,color=black, fill=black, inner sep=0pt,minimum size=2pt,label=below right:{}] (b) at ($(pointer) + (0, -\xx)$) {};
}

\end{tikzpicture}
}
\caption{Our spectral-spatial deep network architecture is divided into the 3D convolutional block which extracts features, and the classification block which performs the final classification. We present the dimensionality of the input patch, initially of size $(\Patchx\times\Patchy\times\Patchw)$, where $\Patchw=\NumberOfBands$, while passing through the consecutive layers of the network. In each feature-extraction layer, we generate 24 feature maps capturing spectral-spatial features, and being the sums of feature maps obtained for each kernel in the corresponding layer. After the last 3D convolutional layer, we flatten all features to fed them into the dense block containing three hidden layers with 512, 256, and 128 neurons, and the final layer with $\NumberOfClasses$ neurons, where $\NumberOfClasses$ equals the number of classes.}\label{fig:network}
\end{figure*}
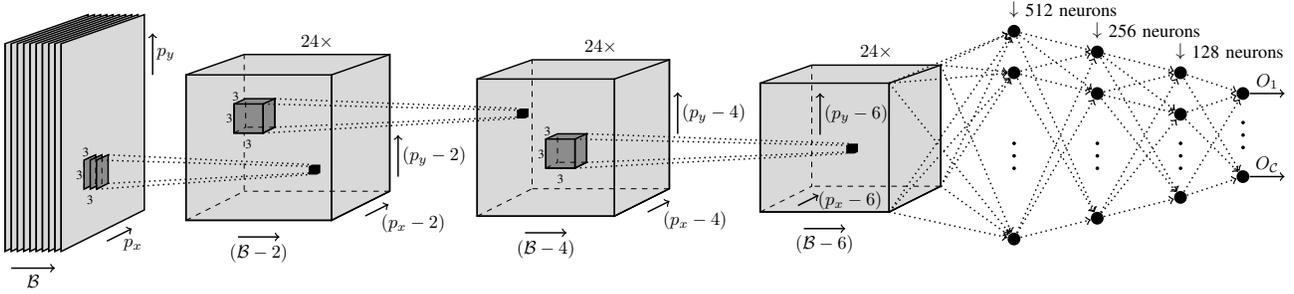

In~\cite{Chen2016}, the authors compared spectral, spatial, and spectral-spatial deep networks, showing that integrating the spectral and spatial features by kernels with large spectral and small spatial receptive fields helps boost the classification accuracy, if the appropriate regularization is applied (due to the limited size of training sets). Spectral-spatial residual networks employ residual blocks to additionally connect every other 3D convolutional layer, and were shown to be alleviating the declining-accuracy phenomenon of other deep learning models~\cite{8061020}. Hamida et al.~\cite{8344565} generate 3D feature maps using 3D kernels which are gradually turned into 1D feature vectors all along the layers in their 3D network. In~\cite{PAOLETTI2018120}, the authors enhanced their spectral-spatial networks with a border mirroring strategy to effectively process the areas which are positioned next to the border of an input HSI. An interesting approach of coupling deep learning with multiple feature learning was introduced in~\cite{Gao_2018}, where a deep network benefits from the additionally-extracted features. Finally, an end-to-end architecture (called BASS) for efficient band-specific feature learning, while keeping the number of trainable parameters as low as possible, was introduced by Santara et al.~\cite{Santara2017}.

Although acquiring hyperspectral data has been becoming much more affordable, its efficient transfer and analysis is still very time- and cost-inefficient. Therefore, generating new ground-truth datasets is infeasible at a large scale. It is reflected in a very small number of hyperspectral benchmarks which may be exploited to train classification engines in a supervised way~\cite{Nalepa2019GRSL}. Overall, there are three approaches for dealing with the problem of limited ground-truth HSI. In \emph{unsupervised segmentation}, prior class labels are not utilized at all, and the input data undergoes clustering into coherent regions of pixels manifesting similar characteristics. The class labels for those regions are, however, unknown, hence they require further post-processing to understand the segmentation results. The unsupervised HSI classification techniques include both standard clustering, performed over the input hyperspectral data or extracted features, and various deep network architectures which perform additional representation learning for more robust classification~\cite{8082108}. In our recent work~\cite{Nalepa2019UnsupervisedHSI}, we introduced an end-to-end approach which combines 3D convolutional autoencoders with a clustering layer for this task.

On the other hand, \emph{transfer learning} and \emph{data augmentation} techniques are appropriate in scenarios in which the labeled training data exists and can be used for training or fine-tuning the pre-trained models. Transfer learning is aimed at pre-training a network using a source dataset (as large and representative as possible), and fine-tune it over the target set (often much smaller) which exposes the characteristics of the test data. In contrast, data augmentation techniques either generate additional training data samples that follow the original data distribution and include them in the training set to enhance its representativeness and size, or elaborate new samples during the inference to benefit from an ensemble-like classification engine~\cite{8746168}. Although data augmentation has been shown very effective in allowing large-capacity learners train from small training samples, it has not been intensively applied in HSI analysis, especially in the case of spectral-spatial deep networks (instead, active learning-guided classification model was applied in this context~\cite{8390931}). Here, we shed more light on the training-time data augmentation for spectral-spatial deep convolutional neural networks, and thoroughly verify its impact on the classification abilities of the models, as well as their training and inference times.

In this letter, we introduce a new spectral-spatial convolutional neural network which combines both spectral and spatial hyperspectral pixel features extracted by its initial 3D convolutional layers (Section~\ref{sec:method}). To fully benefit from the local features in the spectral and spatial dimensions (as subtle differences across materials are often manifested in a tiny part of the spectrum, and the pixel sizes can easily become very large, especially in Earth observation scenarios where HSI is acquired by an imaging satellite), our kernels are kept very small and move with unit stride in each dimension to effectively capture fine-grained features. It is in contrast to other state-of-the-art spectral-spatial models which often employ ``large'' kernels (in the spectral dimension). Our deep network is enhanced by a battery of data augmentation operations applied over 3D hyperspectral patches. A rigorous experimental study performed over the most popular HSI scenes revealed that the proposed approach is competitive against other 3D deep networks from the literature, and offers instant HSI classification (Section~\ref{sec:experiments}). Also, we analyze the statistical impact of training-time data augmentation on the overall performance of such large-capacity learners.

\section{Our Method}\label{sec:method}

We introduce a new spectral-spatial convolutional neural network (Fig.~\ref{fig:network}) which is divided into two parts---the \emph{{\rm 3D} convolutional block}, built with a number of 3D convolutional layers, and the \emph{classification block} performing the final classification of an incoming HSI pixel $\Pixel$. For each $\Pixel$, we extract a surrounding patch $\Patch$ of size $(\Patchx\times\Patchy\times\Patchw)$, where $\Patchx$, $\Patchy$, and $\Patchw=\NumberOfBands$ represent its width, height, and depth (equal to the number of bands in the input HSI), respectively, with $\Pixel$ positioned in its center. To capture fine-grained spectral-spatial features, we apply $\NumberOfKernels=24$ kernels of size $(3\times 3\times 3)$ in each layer, and the patch size of $(7\times 7)$, therefore $\Patchx=\Patchy=7$. These kernels move with unit stride in all dimensions. In the feature extraction part of the network, we exploit zero padding and the rectified linear unit (ReLU) as a non-linearity. Therefore, the dimensionality of the input tensor changes in the consecutive layers---let us assume we want to obtain a class label for a hyperspectral pixel represented by $\mathcal{B}=200$ reflectance values (bands). In this scenario, we would obtain the following tensors corresponding to the feature maps throughout the feature extraction process (starting from the input patch $\Patch$ extracted for this pixel from the input HSI, to the output of the last layer in the 3D convolutional block): $\left[7,7,200\right]\rightarrow\left[5,5,198\right]\rightarrow\left[3,3,196\right]\rightarrow\left[1,1,194\right]$. The number of feature maps at each level remains constant, and equals the number of kernels $\NumberOfKernels=24$. In the inner layers, all 3D kernels ($\NumberOfKernels=24$) are convolved with each feature map, and the resulting intermediate feature maps ($\NumberOfKernels$ for each kernel) are summed together to form the final map for a specific kernel. In the classification part of the model, a dense block with four fully-connected layers (with ReLU activation) containing 512, 256, 128, and $\NumberOfClasses$ neurons, where $\NumberOfClasses$ denotes the number of classes in the corresponding dataset, is exploited.

\newcommand{\mybackground}{35}
\begin{table*}[ht!]
	\scriptsize
	\centering
	\caption{The results obtained using all variants of the investigated methods for Salinas Valley.}
	\label{tab:salinas_valley}
	\renewcommand{\tabcolsep}{0.15cm}
	\begin{tabular}{r|cccccccccccccccc|cc}

\hline
Algorithm& C1 & C2 & C3 & C4 & C5 & C6 & C7 & C8 & C9 & C10 & C11 & C12 & C13 & C14 & C15 & C16 & OA & AA\\
\hline
3D-CNN&\cellcolor{green!40}96.49&75.15&39.89&61.61&51.78&79.21&76.81&74.84&78.14&85.69&\cellcolor{green!40}71.56&76.49&\cellcolor{gray!\mybackground}80.86&62.15&61.80&33.00&69.72&69.09\\
3D-CNN-Rotate&\cellcolor{gray!\mybackground}90.83&\cellcolor{green!40}85.00&39.64&74.26&\cellcolor{green!40}54.51&79.44&78.42&\cellcolor{green!40}\textbf{78.00}&76.91&84.55&67.96&\cellcolor{green!40}79.83&\cellcolor{green!40}97.56&\cellcolor{green!40}75.40&\cellcolor{green!40}62.09&\cellcolor{green!40}34.62&\cellcolor{green!40}72.01&\cellcolor{green!40}72.44\\
3D-CNN-Flip&94.63&80.41&\cellcolor{red!30}38.85&72.09&\cellcolor{gray!\mybackground}49.43&79.26&78.11&76.09&\cellcolor{red!30}76.01&\cellcolor{red!30}83.15&\cellcolor{red!30}62.62&79.61&85.03&53.64&\cellcolor{red!30}55.10&32.64&69.04&68.54\\
3D-CNN-Zoom&91.55&\cellcolor{red!30}71.35&38.89&\cellcolor{red!30}58.61&53.01&\cellcolor{green!40}79.60&\cellcolor{red!30}76.37&\cellcolor{gray!\mybackground}73.75&76.82&85.67&69.19&\cellcolor{red!30}73.69&89.45&\cellcolor{red!30}53.44&60.41&33.34&\cellcolor{red!30}68.47&\cellcolor{red!30}67.82\\
3D-CNN-Mixed&95.12&80.70&\cellcolor{green!40}40.79&\cellcolor{green!40}74.82&53.36&\cellcolor{gray!\mybackground}79.20&\cellcolor{green!40}78.99&75.77&\cellcolor{green!40}79.50&\cellcolor{green!40}86.48&64.53&79.73&97.16&67.40&60.76&\cellcolor{gray!\mybackground}32.43&71.38&71.67\\
\hline
{BASS}&97.78&85.01&\cellcolor{green!40}\textbf{59.30}&\cellcolor{gray!\mybackground}78.70&49.23&\cellcolor{red!30}79.05&\cellcolor{gray!\mybackground}76.76&\cellcolor{red!30}70.62&\cellcolor{green!40}86.31&\cellcolor{gray!\mybackground}83.47&73.64&\cellcolor{gray!\mybackground}77.30&\cellcolor{red!30}68.50&\cellcolor{gray!\mybackground}66.41&58.03&31.39&\cellcolor{gray!\mybackground}70.93&\cellcolor{gray!\mybackground}71.35\\
{BASS-Rotate}&97.65&\cellcolor{green!40}90.12&53.04&\cellcolor{green!40}\textbf{79.27}&\cellcolor{green!40}52.98&\cellcolor{green!40}\textbf{79.93}&\cellcolor{green!40}79.23&\cellcolor{green!40}75.99&83.32&\cellcolor{green!40}88.21&\cellcolor{gray!\mybackground}72.85&\cellcolor{green!40}82.05&\cellcolor{green!40}85.94&75.51&58.61&\cellcolor{green!40}34.27&\cellcolor{green!40}73.38&\cellcolor{green!40}74.31\\
{BASS-Flip}&\cellcolor{green!40}\textbf{98.49}&\cellcolor{gray!\mybackground}83.28&53.02&78.96&\cellcolor{red!30}48.12&79.78&78.48&75.06&\cellcolor{gray!\mybackground}81.84&85.08&\cellcolor{green!40}74.38&80.56&74.70&76.43&\cellcolor{gray!\mybackground}55.41&\cellcolor{red!30}30.90&71.29&72.16\\
{BASS-Zoom}&97.40&87.74&50.79&78.82&52.80&79.36&78.18&74.43&83.11&84.87&73.68&78.55&69.79&67.46&57.07&34.02&71.62&71.75\\
{BASS-Mixed}&\cellcolor{gray!\mybackground}95.42&88.22&\cellcolor{gray!\mybackground}50.15&78.87&50.13&79.68&79.01&75.16&85.28&86.54&74.15&81.92&83.26&\cellcolor{green!40}78.87&\cellcolor{green!40}61.77&32.21&73.28&73.79\\
\hline
\textbf{Ours}&93.22&\cellcolor{green!40}\textbf{90.48}&47.57&78.97&57.35&\cellcolor{gray!\mybackground}79.48&79.13&75.77&\cellcolor{gray!\mybackground}79.71&\cellcolor{green!40}\textbf{88.59}&74.68&\cellcolor{gray!\mybackground}83.48&94.53&\cellcolor{gray!\mybackground}85.36&\cellcolor{gray!\mybackground}62.52&\cellcolor{gray!\mybackground}32.47&\cellcolor{gray!\mybackground}73.72&\cellcolor{gray!\mybackground}75.21\\
\textbf{Ours-Rotate}&95.30&90.30&\cellcolor{green!40}50.35&78.89&56.59&79.56&\cellcolor{green!40}\textbf{79.38}&76.37&\cellcolor{green!40}\textbf{86.89}&88.01&74.79&84.77&98.55&\cellcolor{green!40}\textbf{87.76}&\cellcolor{green!40}\textbf{65.67}&\cellcolor{green!40}\textbf{34.97}&\cellcolor{green!40}\textbf{75.54}&\cellcolor{green!40}\textbf{76.76}\\
\textbf{Ours-Flip}&\cellcolor{green!40}95.95&89.58&47.63&\cellcolor{green!40}78.98&\cellcolor{green!40}\textbf{57.87}&79.64&79.19&\cellcolor{green!40}76.64&84.50&\cellcolor{gray!\mybackground}87.92&\cellcolor{gray!\mybackground}73.79&\cellcolor{green!40}\textbf{85.18}&\cellcolor{gray!\mybackground}94.29&86.54&62.94&34.79&74.77&75.96\\
\textbf{Ours-Zoom}&95.65&\cellcolor{gray!\mybackground}81.47&47.61&\cellcolor{gray!\mybackground}78.54&\cellcolor{gray!\mybackground}55.22&\cellcolor{green!40}79.65&\cellcolor{gray!\mybackground}78.66&\cellcolor{gray!\mybackground}75.73&86.61&88.24&\cellcolor{green!40}\textbf{76.37}&85.14&\cellcolor{green!40}\textbf{98.94}&85.41&62.68&33.36&74.04&75.58\\
\textbf{Ours-Mixed}&\cellcolor{red!30}89.95&85.48&\cellcolor{gray!\mybackground}47.25&78.78&57.20&79.64&79.30&76.36&83.51&88.23&74.91&84.09&97.83&85.56&63.67&34.83&74.13&75.41\\
\hline
\multicolumn{19}{c}{\scriptsize \textbf{How to read this table:} The \emph{globally} best result (across all techniques) is boldfaced, and the background of the globally worst cell is red. }\\
\multicolumn{19}{c}{\scriptsize For \emph{each method} (3D-CNN, BASS, and our deep network) separately, the background of the best variant is green, and the background of the worst variant is gray.}\\
	\end{tabular}
\end{table*}

\begin{table*}[ht!]
	\scriptsize
	\centering
	\caption{The results obtained using all variants of the investigated methods for Indian Pines. The meanings of colors as in Table~\ref{tab:salinas_valley}.}
	\label{tab:indian_pines}
	\renewcommand{\tabcolsep}{0.15cm}
	\begin{tabular}{r|cccccccccccccccc|cc}
\hline
Algorithm& C1 & C2 & C3 & C4 & C5 & C6 & C7 & C8 & C9 & C10 & C11 & C12 & C13 & C14 & C15 & C16 & OA & AA\\
\hline
3D-CNN&\cellcolor{gray!\mybackground}5.00&\cellcolor{gray!\mybackground}33.70&\cellcolor{gray!\mybackground}28.30&\cellcolor{red!30}17.88&51.32&\cellcolor{gray!\mybackground}60.18&\cellcolor{red!30}0.00&\cellcolor{red!30}65.99&\cellcolor{gray!\mybackground}1.67&\cellcolor{gray!\mybackground}53.06&\cellcolor{gray!\mybackground}54.27&23.20&65.87&\cellcolor{red!30}77.01&\cellcolor{green!40}37.95&\cellcolor{gray!\mybackground}37.94&\cellcolor{gray!\mybackground}48.89&\cellcolor{gray!\mybackground}38.33\\
3D-CNN-Rotate&6.25&\cellcolor{green!40}49.53&\cellcolor{green!40}\textbf{46.13}&\cellcolor{green!40}\textbf{44.73}&\cellcolor{green!40}65.92&\cellcolor{green!40}90.42&\cellcolor{red!30}0.00&\cellcolor{green!40}\textbf{84.73}&\cellcolor{green!40}\textbf{26.53}&\cellcolor{green!40}\textbf{68.79}&60.12&\cellcolor{green!40}27.87&\cellcolor{green!40}84.07&79.79&36.45&43.85&\cellcolor{green!40}60.34&\cellcolor{green!40}\textbf{50.95}\\
3D-CNN-Flip&8.75&42.18&31.59&36.62&52.25&81.79&\cellcolor{red!30}0.00&82.05&20.83&62.69&60.96&23.40&71.69&82.66&33.85&\cellcolor{green!40}\textbf{49.67}&55.98&46.31\\
3D-CNN-Zoom&5.18&36.19&29.73&23.48&\cellcolor{gray!\mybackground}51.18&81.86&\cellcolor{red!30}0.00&76.07&25.21&58.19&\cellcolor{green!40}\textbf{67.88}&\cellcolor{gray!\mybackground}21.52&\cellcolor{gray!\mybackground}65.50&\cellcolor{green!40}84.54&\cellcolor{gray!\mybackground}23.56&42.83&55.25&43.31\\
3D-CNN-Mixed&\cellcolor{green!40}\textbf{12.32}&45.18&40.46&40.18&62.46&84.71&\cellcolor{red!30}0.00&80.76&26.42&66.34&60.37&27.50&80.85&82.06&36.36&49.18&58.52&50.44\\
\hline
{BASS}&\cellcolor{red!30}0.00&29.68&16.49&29.45&\cellcolor{red!30}22.58&59.39&\cellcolor{red!30}0.00&80.43&\cellcolor{red!30}0.00&26.99&\cellcolor{green!40}65.47&\cellcolor{red!30}16.28&\cellcolor{red!30}15.71&82.32&21.61&\cellcolor{red!30}23.89&45.30&\cellcolor{red!30}30.64\\
{BASS-Rotate}&1.00&\cellcolor{green!40}40.66&\cellcolor{red!30}16.38&32.87&\cellcolor{green!40}30.95&\cellcolor{green!40}70.55&\cellcolor{red!30}0.00&\cellcolor{green!40}82.57&\cellcolor{red!30}0.00&41.57&51.77&\cellcolor{green!40}20.63&27.02&83.85&\cellcolor{red!30}21.36&34.14&\cellcolor{green!40}46.82&34.69\\
{BASS-Flip}&1.00&36.08&19.38&26.70&24.13&67.54&\cellcolor{red!30}0.00&78.72&\cellcolor{red!30}0.00&\cellcolor{red!30}36.92&55.56&17.21&27.20&\cellcolor{green!40}85.31&27.03&32.73&46.38&33.45\\
{BASS-Zoom}&\cellcolor{red!30}0.00&32.31&23.54&\cellcolor{gray!\mybackground}25.27&28.31&\cellcolor{red!30}57.17&\cellcolor{red!30}0.00&82.01&\cellcolor{red!30}0.00&41.74&\cellcolor{red!30}50.21&18.26&20.65&\cellcolor{gray!\mybackground}81.97&22.43&27.82&\cellcolor{red!30}44.35&31.98\\
{BASS-Mixed}&\cellcolor{green!40}1.61&\cellcolor{red!30}29.37&\cellcolor{green!40}25.23&\cellcolor{green!40}33.46&27.81&61.32&\cellcolor{red!30}0.00&\cellcolor{gray!\mybackground}78.71&\cellcolor{red!30}0.00&\cellcolor{green!40}47.85&53.19&18.34&\cellcolor{green!40}37.40&83.24&\cellcolor{green!40}28.71&\cellcolor{green!40}40.80&46.33&\cellcolor{green!40}35.44\\
\hline
\textbf{Ours}&\cellcolor{green!40}5.18&43.14&\cellcolor{gray!\mybackground}27.72&24.74&\cellcolor{gray!\mybackground}56.02&\cellcolor{gray!\mybackground}83.89&\cellcolor{red!30}0.00&72.59&5.85&\cellcolor{gray!\mybackground}53.12&65.39&20.35&\cellcolor{gray!\mybackground}55.49&81.00&33.74&35.22&\cellcolor{gray!\mybackground}54.93&41.47\\
\textbf{Ours-Rotate}&\cellcolor{green!40}5.18&\cellcolor{green!40}\textbf{52.80}&\cellcolor{green!40}38.00&40.58&\cellcolor{green!40}\textbf{71.54}&\cellcolor{green!40}\textbf{93.65}&\cellcolor{red!30}0.00&69.96&\cellcolor{green!40}10.56&\cellcolor{green!40}63.07&65.76&\cellcolor{green!40}\textbf{29.44}&\cellcolor{green!40}\textbf{89.07}&83.00&\cellcolor{green!40}\textbf{40.92}&\cellcolor{green!40}40.00&\cellcolor{green!40}\textbf{61.33}&\cellcolor{green!40}49.60\\
\textbf{Ours-Flip}&4.46&43.54&32.76&37.43&60.21&87.57&\cellcolor{red!30}0.00&\cellcolor{gray!\mybackground}66.43&3.33&59.62&\cellcolor{gray!\mybackground}62.64&22.13&56.23&82.08&36.54&34.73&56.06&43.11\\
\textbf{Ours-Zoom}&\cellcolor{gray!\mybackground}1.00&\cellcolor{gray!\mybackground}40.42&29.22&\cellcolor{gray!\mybackground}21.57&58.49&84.53&\cellcolor{red!30}0.00&70.48&\cellcolor{gray!\mybackground}1.00&59.43&\cellcolor{green!40}67.34&\cellcolor{gray!\mybackground}20.00&60.32&\cellcolor{gray!\mybackground}80.55&\cellcolor{gray!\mybackground}29.36&\cellcolor{gray!\mybackground}33.88&55.46&\cellcolor{gray!\mybackground}41.07\\
\textbf{Ours-Mixed}&1.25&51.55&35.65&\cellcolor{green!40}42.43&67.20&90.43&\cellcolor{red!30}0.00&\cellcolor{green!40}74.90&5.56&61.10&64.60&24.38&77.20&\cellcolor{green!40}\textbf{86.00}&34.98&38.23&60.01&47.22\\
\hline
	\end{tabular}
\end{table*}

\begin{table*}[ht!]
	\scriptsize
	\centering
	\caption{The results obtained using all variants of the investigated methods for Pavia University. The meanings of colors as in Table~\ref{tab:salinas_valley}.}
	\label{tab:pavia_university}
	\renewcommand{\tabcolsep}{0.4cm}
	\begin{tabular}{r|ccccccccc|cc}
\hline
Algorithm& C1 & C2 & C3 & C4 & C5 & C6 & C7 & C8 & C9 & OA & AA\\
\hline
3D-CNN&90.66&81.85&41.92&93.02&\cellcolor{gray!\mybackground}59.79&25.20&\cellcolor{red!30}0.00&\cellcolor{red!30}70.18&\cellcolor{gray!\mybackground}79.03&\cellcolor{gray!\mybackground}70.07&60.18\\
3D-CNN-Rotate&\cellcolor{green!40}92.96&82.39&41.96&93.46&59.93&\cellcolor{green!40}\textbf{26.38}&\cellcolor{red!30}0.00&73.98&\cellcolor{green!40}\textbf{79.89}&\cellcolor{green!40}\textbf{71.19}&\cellcolor{green!40}\textbf{62.93}\\
3D-CNN-Flip&91.15&82.45&\cellcolor{gray!\mybackground}39.63&92.57&59.97&25.44&\cellcolor{red!30}0.00&\cellcolor{green!40}75.37&79.40&70.77&60.66\\
3D-CNN-Zoom&\cellcolor{red!30}89.12&\cellcolor{green!40}\textbf{83.18}&41.09&\cellcolor{gray!\mybackground}92.34&59.91&\cellcolor{gray!\mybackground}25.01&\cellcolor{red!30}0.00&70.82&79.32&70.37&\cellcolor{gray!\mybackground}60.09\\
3D-CNN-Mixed&92.45&\cellcolor{gray!\mybackground}81.49&\cellcolor{green!40}44.07&\cellcolor{green!40}\textbf{94.13}&\cellcolor{green!40}\textbf{59.99}&26.14&\cellcolor{red!30}0.00&74.52&79.39&70.88&61.35\\
\hline
{BASS}&91.27&79.34&33.96&90.47&59.86&24.45&\cellcolor{red!30}0.00&72.94&77.96&68.62&58.92\\
{BASS-Rotate}&\cellcolor{green!40}93.55&78.49&\cellcolor{red!30}30.43&92.29&59.96&25.21&\cellcolor{red!30}0.00&81.51&\cellcolor{green!40}79.26&69.44&60.01\\
{BASS-Flip}&\cellcolor{gray!\mybackground}91.13&\cellcolor{green!40}82.13&\cellcolor{green!40}41.07&\cellcolor{red!30}88.84&\cellcolor{red!30}49.83&\cellcolor{red!30}17.00&\cellcolor{red!30}0.00&74.87&\cellcolor{red!30}72.49&68.82&\cellcolor{red!30}57.48\\
{BASS-Zoom}&92.90&\cellcolor{gray!\mybackground}77.96&36.55&91.53&\cellcolor{green!40}\textbf{59.99}&25.16&\cellcolor{red!30}0.00&\cellcolor{gray!\mybackground}71.49&77.31&\cellcolor{red!30}68.41&59.21\\
{BASS-Mixed}&92.52&78.25&34.61&\cellcolor{green!40}92.47&59.82&\cellcolor{green!40}25.35&\cellcolor{red!30}0.00&\cellcolor{green!40}\textbf{82.54}&78.18&\cellcolor{green!40}69.48&\cellcolor{green!40}60.42\\
\hline
\textbf{Ours}&94.13&\cellcolor{green!40}77.17&\cellcolor{gray!\mybackground}43.38&92.37&59.90&\cellcolor{green!40}23.52&\cellcolor{red!30}0.00&\cellcolor{gray!\mybackground}75.79&77.47&68.83&\cellcolor{gray!\mybackground}60.41\\
\textbf{Ours-Rotate}&\cellcolor{green!40}\textbf{95.76}&76.78&45.86&\cellcolor{green!40}93.58&\cellcolor{gray!\mybackground}59.89&\cellcolor{gray!\mybackground}22.83&\cellcolor{red!30}0.00&76.23&\cellcolor{green!40}78.22&69.11&61.02\\
\textbf{Ours-Flip}&\cellcolor{gray!\mybackground}93.96&\cellcolor{red!30}75.61&45.93&92.47&59.93&22.90&\cellcolor{red!30}0.00&79.93&77.03&\cellcolor{gray!\mybackground}68.56&60.86\\
\textbf{Ours-Zoom}&94.04&76.98&\cellcolor{green!40}\textbf{46.20}&\cellcolor{gray!\mybackground}91.88&59.96&22.60&\cellcolor{red!30}0.00&76.11&\cellcolor{gray!\mybackground}76.74&68.76&60.50\\
\textbf{Ours-Mixed}&94.79&77.05&45.43&92.87&\cellcolor{green!40}59.97&23.35&\cellcolor{red!30}0.00&\cellcolor{green!40}80.57&77.88&\cellcolor{green!40}69.44&\cellcolor{green!40}61.32\\
\hline
	\end{tabular}
\end{table*}

Since the capacity of the model becomes very large (it could be decreased by applying a shallower dense block or other supervised learner though), and the network can easily overfit to the training data, we augment the training sets and generate synthetic training patches using rotation, flipping, zooming, and a random mix of all of the aforementioned operations, each applied with the same probability. Since the augmentations are performed at the \emph{patch level}, we extract larger patches for rotation, to make sure that the rotated patches are not cropped (and to ultimately avoid padding). Such artificial patches are appended to the training set---we at most doubled the number of original samples, unless that number would exceed the most numerous class. In this case, we augment only the missing difference, to ensure that at least half of the training examples are original.

Training of the deep network proceeds until the termination condition has been reached: the maximum execution time has elapsed, the number of epochs has been processed, or an early stopping has been triggered (e.g.,~the loss value stagnates for a given number of consecutive epochs). In this letter, we combine the early stopping (15 training epochs without improvement in the cross-entropy loss value) with the maximum number of epochs (in all experiments reported in this work, we restricted the number of epochs to 200).

\section{Experiments}\label{sec:experiments}

The objectives of our experimental study are two-fold. First, we compare our 3D convolutional neural network with two other spectral-spatial techniques from the literature: the BASS deep network~\cite{Santara2017}, and a method coupling deep networks with multiple feature learning (we refer to it as 3D-CNN)~\cite{Gao_2018}. Second, we investigate the impact of our training-time data augmentation procedures on the abilities of the deep models. For fair comparison, we execute all of our training-time data augmentations for both deep models from the literature too.

In this letter, we exploited three most popular HSI benchmarks\footnote{These datasets are available at~\url{http://www.ehu.eus/ccwintco/index.php/Hyperspectral_Remote_Sensing_Scenes}.} from the literature~\cite{Nalepa2019GRSL}: (i)~Salinas Valley (Sa), USA ($217\times 512$ pixels, acquired by the AVIRIS sensor) presenting different sorts of vegetation (16 classes, 224 bands, with 3.7~m spatial resolution); (ii)~Indian Pines (IP), USA ($145\times 145$, AVIRIS) covering the agriculture and forest (16 classes, 200 channels, 20~m); (iii)~Pavia University (PU), Italy ($340\times 610$, ROSIS) capturing the urban scenery (9 classes, 103 channels, 1.3 m). We report the average accuracy (AA), overall accuracy (OA), and the kappa scores $\kappa=1-\frac{1-p_o}{1-p_e}$, where $p_o$ and $p_e$ are the relative observed agreement, and hypothetical probability of chance agreement, respectively, and $-1\leq \kappa\leq 1$ (the larger $\kappa$ is, the better performance is), obtained over the test sets. To avoid any training-test information leaks, we utilize our challenging patch-based data splits~\cite{Nalepa2019GRSL}, and perform five-fold cross-validation\footnote{Our patch-based data splits are available at \url{https://tinyurl.com/ieee-grsl}.} (we run the algorithms $5\times$ for each fold, thus we have 25 executions of each method altogether). The deep neural network models were implemented in \texttt{Python 3.6}, and were trained using Adam with the learning rate of $10^{-4}$, $\beta_1 = 0.9$, $\beta_2 = 0.999$, and the batch size of 64.

The experimental results are gathered in Tables~\ref{tab:salinas_valley}--\ref{tab:pavia_university} for Salinas Valley, Indian Pines, and Pavia University, respectively. We can appreciate that our deep models lead to consistently high-quality classification for all datasets, and significantly outperform BASS for all sets, and 3D-CNN for Salinas Valley and Indian Pines. Importantly, there are only sporadic cases in which our method delivered the globally worst results for specific classes (C1 for Sa, and C2 for PU obtained with our deep network coupled with the mixed and flip augmentation, respectively). It indicates that the proposed architecture deals well with very imbalanced and under-represented classes, and offers accurate classification in such scenarios. Note that for C7 in IP and PU all investigated methods failed, and obtained 0\% accuracy---for these classes, the patch-based division contained either no examples in the training or test sets, hence the models were not able to effectively learn those classes (therefore, they could not appropriately cope with unknown examples presented in the test sets of the corresponding folds).

\begin{figure}[ht!]
\begin{tabular}{c}
  \includegraphics[width=0.9\columnwidth]{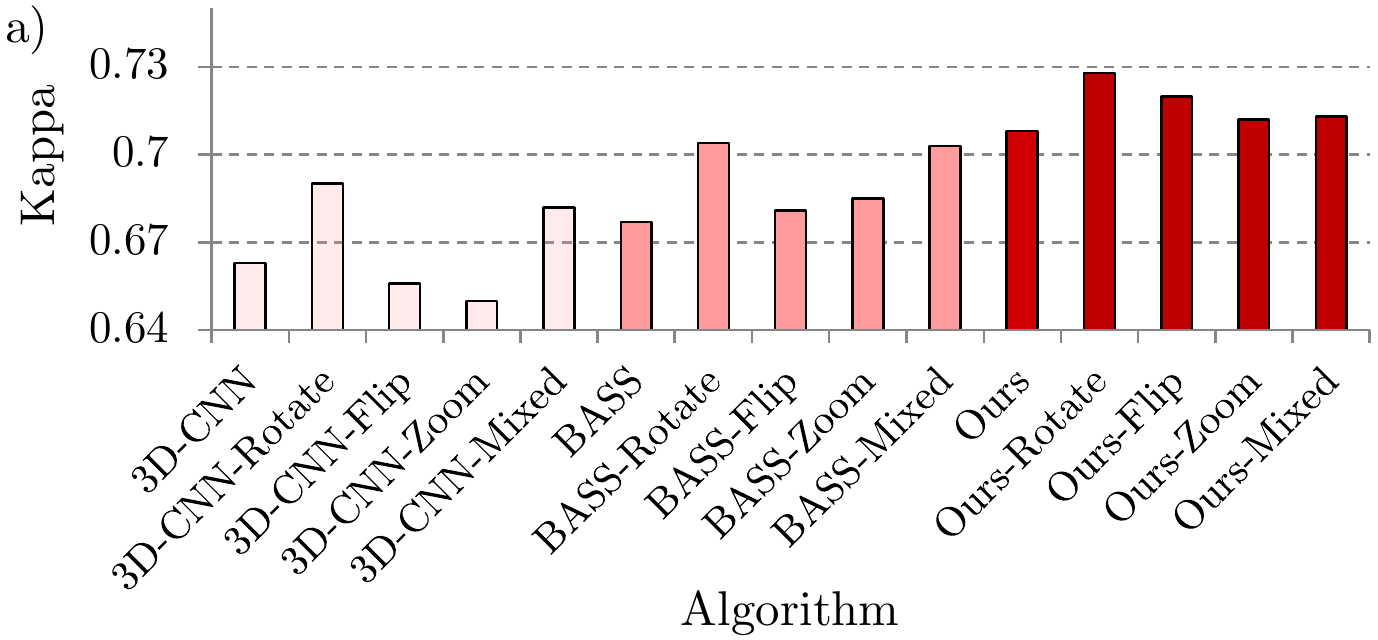}\\
  \includegraphics[width=0.9\columnwidth]{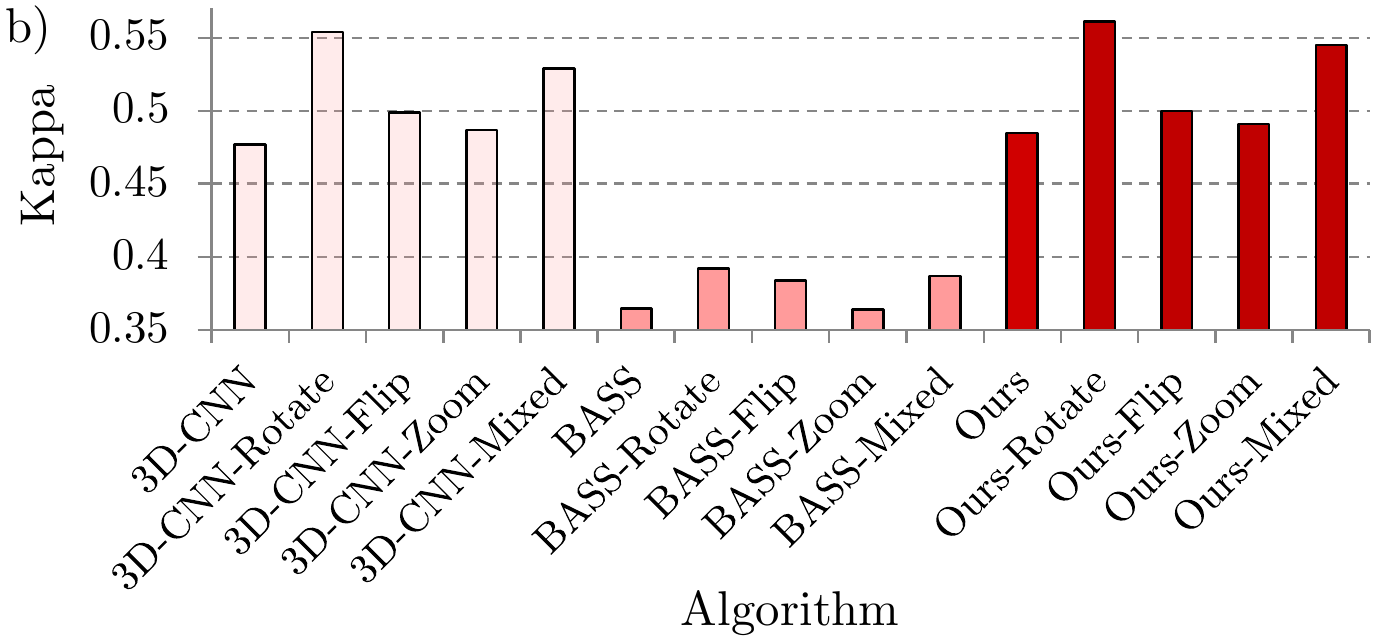}\\
  \includegraphics[width=0.9\columnwidth]{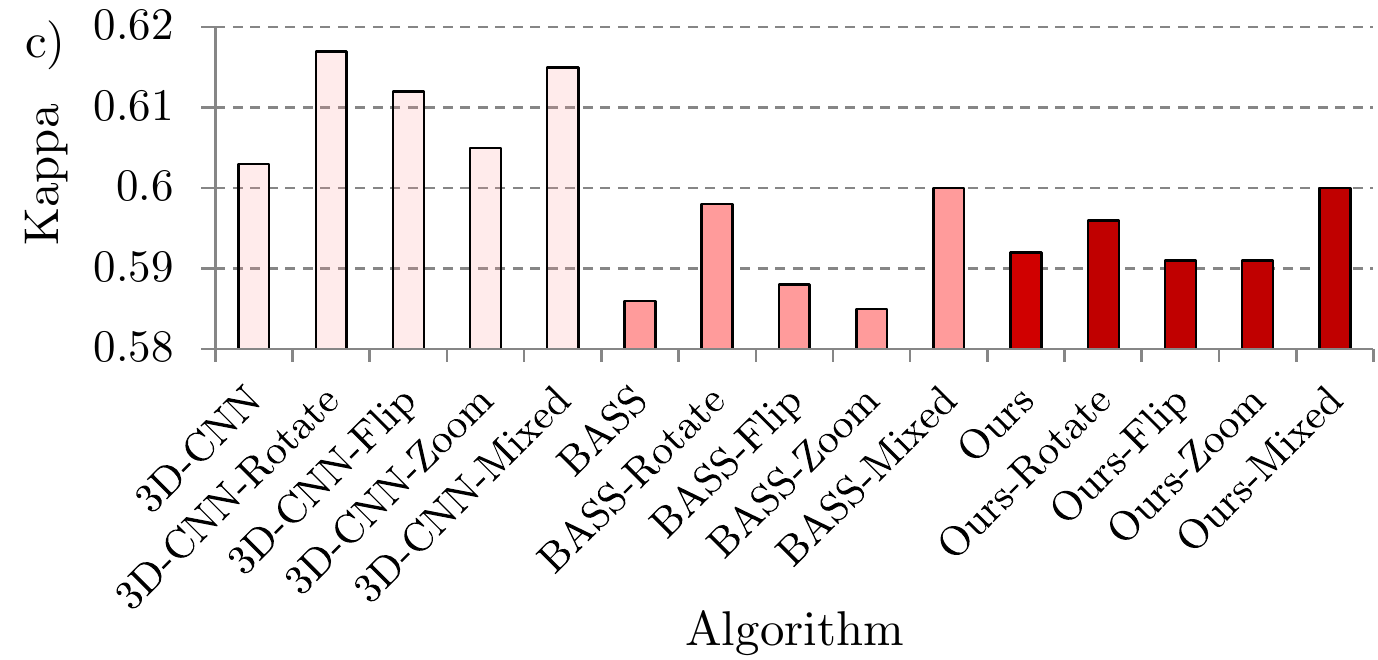}\\
\end{tabular}
	\caption[fig:kappa_Sa]{The kappa scores obtained using all methods for all datasets: a)~Salinas Valley, b)~Indian Pines, and c)~Pavia University.}
	\label{fig:kappa_Sa}
\end{figure}

In Fig.~\ref{fig:kappa_Sa}, we visualize the average kappa scores obtained for all deep models in all investigated augmentation scenarios. The results confirm our previous observations---our spectral-spatial networks obtained better classification when compared with BASS over all training-time augmentations. On the other hand, learning additional features which is a pivotal part of 3D-CNN appeared extremely useful in Pavia University, where 3D-CNN outperformed both BASS and our deep models by a significant margin---see the pair-wise comparison of 3D-CNN with BASS and Ours in Fig.~\ref{fig:kappa_Sa}c. It indicates that for the sets in which there exist over-represented classes (the Meadows class constitutes more than 40\% of all samples) extracting additional features, exploited later by deep models, is extremely beneficial, as the deep nets are unable to capture distinctive features of the minority classes from very limited training samples.

The ranking obtained over the kappa scores, averaged across all hyperspectral scenes (AR) shows that our models delivered better classification results when compared with both BASS and 3D-CNN on average (Table~\ref{tab:times}). Also, training-time data augmentation brought noticeable improvements in classification accuracy for all networks, with the data augmentation exploiting only rotation being the best for all deep learners.

\begin{table}[ht!]
	\scriptsize
	\centering
	\caption{The ranking of all methods (averaged across all sets; AR), alongside the training times~(s), and the inference time of classifying \emph{one} test sample~(ms). The best ranking is boldfaced.}
	\label{tab:times}
	\renewcommand{\tabcolsep}{0.15cm}
	\begin{tabular}{r|r|rrr|rrr}
		\hline
&  & \multicolumn{3}{c|}{Training time} & \multicolumn{3}{c}{Inference time} \\
\hline
Algorithm$\downarrow$ & AR & Sa & IP & PU & Sa & IP & PU\\
\hline
3D-CNN        & 9.33 & 77.38 & 31.38 & 19.23 & 0.17 & 0.08& 0.08\\
3D-CNN-Rotate     & \textbf{3.67} & 112.99 & 46.94 & 32.34 &0.17 &0.08 & 0.08\\
3D-CNN-Flip       & 7.67 & 125.84& 47.96 & 30.72 & 0.17& 0.09& 0.08\\
3D-CNN-Zoom       & 9.00 & 102.23 & 42.54 & 34.36 & 0.17&0.08 &0.08 \\
3D-CNN-Mixed      & 5.33 & 125.81& 44.35 & 30.39  & 0.17&0.08 &0.08 \\
\hdashline
{BASS} & 13.33 & 339.60 & 207.64 & 75.01 & 0.29 & 0.27  & 0.12 \\
{BASS-Rotate} & 8.33 & 585.20 & 241.12  & 85.78 & 0.29 & 0.27  & 0.12 \\
{BASS-Flip} & 12.33 & 586.21 & 346.53 & 119.61 & 0.29 & 0.27  & 0.12 \\
{BASS-Zoom} & 13.00 & 619.48  & 339.39  & 81.54 & 0.29  & 0.27  & 0.12 \\
{BASS-Mixed} & 8.50 & 552.16 & 316.34 & 100.68 & 0.29 & 0.27 & 0.12 \\
\hdashline
\textbf{Ours} & 8.00 & 157.32 & 103.69 & 29.58  & 0.18 & 0.16  & 0.09 \\
\textbf{Ours-Rotate} & \textbf{3.67} & 248.24 & 156.00 & 48.26 & 0.18 & 0.16  & 0.09 \\
\textbf{Ours-Flip} & 6.17 & 255.40 & 161.30 & 40.66 & 0.18 & 0.16  & 0.09 \\
\textbf{Ours-Zoom} & 7.50 & 218.86 & 155.68 & 33.87 & 0.18  & 0.16  & 0.09 \\
\textbf{Ours-Mixed} & 4.17 & 232.53 & 163.47 & 36.12 & 0.18 & 0.16 & 0.09 \\
\hline
	\end{tabular}
\end{table}

To verify the statistical importance of the experimental results, we executed two-tailed Wilcoxon tests over the null hypothesis saying that ``applying 3D-CNN and our spectral-spatial deep models leads to the same-quality per-class classification'' (we focused on these two models as they clearly outperformed the BASS architecture, as presented in Tables~\ref{tab:salinas_valley}--\ref{tab:times} and Fig.~\ref{fig:kappa_Sa}). The results show that this hypothesis can be safely rejected at $p<0.05$ (Table~\ref{tab:wilcoxon_3d_ours})---our deep models delivered statistically better per-class classification accuracies. For details, see the class accuracies for Ours and 3D-CNN reported in Tables~\ref{tab:salinas_valley}--\ref{tab:pavia_university}---they should be confronted in a pair-wise manner, for all augmentation techniques separately.

\begin{table}[ht!]
	\scriptsize
	\centering
	\caption{Wilcoxon tests show that the differences between our deep network and 3D-CNN are statistically important for almost all variants (the $p$-values that show the statistical importance at $p<0.05$ are boldfaced).}
	\label{tab:wilcoxon_3d_ours}
	\renewcommand{\tabcolsep}{0.45cm}
	\begin{tabular}{ccccc}
\hline
Without & Rotate & Flip & Zoom & Mixed\\
\hline
\textbf{$<$0.005}	& \textbf{$<$0.02}	& \textbf{$<$0.02}	& \textbf{$<$0.05}	& {$>$0.1}\\
\hline
	\end{tabular}
\end{table}

In Table~\ref{tab:wilcoxon_data_augmentation}, we present the $p$-values resulting from the Wilcoxon tests verifying the importance of the training-time data augmentation applied to our deep models. Similarly, extending training sets with synthetically-generated examples leads to statistically different classification performance in almost all cases (the augmentation which involved only flipping is an exception here). Also, zooming and flipping of the the training patches gave the same-quality classification. Overall, the results show that applying data augmentation is a powerful remedy in dealing with small training samples for spectral-spatial deep networks, and can significantly boost the abilities of the underlying (large-capacity) deep learners.

\begin{table}[ht!]
	\scriptsize
	\centering
	\caption{Training-time data augmentation applied to the proposed spectral-spatial deep network brings statistically significant improvements in almost all cases (the $p$-values that show the statistical importance at $p<0.05$ are boldfaced). }
	\label{tab:wilcoxon_data_augmentation}
	\renewcommand{\tabcolsep}{0.4cm}
	\begin{tabular}{r|llll}
\hline
 & Rotate & Flip & Zoom & Mixed\\
\hline
Without & $<$\textbf{0.001}	& $>$0.2	& $<$\textbf{0.05}	& $<$\textbf{0.001}\\
Rotate & & $<$\textbf{0.001} & $<$\textbf{0.001} & $<$\textbf{0.05}\\
Flip & & & $>$0.05 & $<$\textbf{0.001}\\
Zoom & & & & $<$\textbf{0.001}\\
\hline
	\end{tabular}
\end{table}

Finally, we analyzed the training and inference times of the investigated techniques in Table~\ref{tab:times}. Since training-time augmentation increases the size of the training set (which is in contrast to the test-time augmentation~\cite{8746168}), these approaches will always affect the training times of any supervised learner. On the other hand, the time of classifying test examples during the operational phase of a learned model is extremely fast in all methods, and all of them offer real-time operation. Also, it is worth noticing that all of the models can be trained in very affordable time with a single GPU---even in the case of the models introduced in this letter, in which a significant number of small 3D convolutional kernels move in both spectral and spatial dimensions while processing an input HSI.

\section{Conclusions}\label{sec:conclusions}

We introduced a new spectral-spatial deep neural network architecture that is composed of the convolutional block which extracts features from the input HSI by applying 3D convolutional layers, and the classification block which performs the final classification. Our rigorous experimental study backed up with statistical tests revealed that the proposed method is competitive with other spectral-spatial architectures from the literature, not only in terms of the classification abilities, but also in terms of training and test time. Although our networks employ small 3D kernels in the convolutional block, the experiments showed that it offers real-time inference. Finally, we showed that applying training-time data augmentation to enhance the representativeness of training sets significantly improves the abilities of underlying deep models.

\ifCLASSOPTIONcaptionsoff
  \newpage
\fi

\bibliographystyle{ieeetran}
\bibliography{IEEEabrv,ref_all}

\end{document}